**Model Tuning or Prompt Tuning? A Study of Large Language Models for Clinical Concept and Relation Extraction**

Authors:   Cheng Peng, PhD[1]
Xi Yang, PhD[1,2]
Kaleb E Smith, PhD[3]
Zehao, Yu, MS[1]
Aokun Chen, PhD[1,2]
Jiang Bian, PhD[1,2]
Yonghui Wu, PhD[1,2]

Affiliation of the authors:   [1]Department of Health Outcomes and Biomedical Informatics, College of Medicine, University of Florida, Gainesville, Florida, USA

[2]Cancer Informatics Shared Resource, University of Florida Health Cancer Center, Gainesville, Florida, USA

[3]NVIDIA, Santa Clara, California, USA

Corresponding author:   Yonghui Wu, PhD
Clinical and Translational Research Building
2004 Mowry Road, PO Box 100177
Gainesville, FL, USA, 32610
Phone: 352-294-8436
Email: yonghui.wu@ufl.edu

Keywords:   Clinical concept extraction

Word count: 4271 (exclude the figures and tables)

**ABSTRACT**

**Objective**

To develop soft prompt-based learning algorithms for large language models (LLMs), examine the shape of prompts, prompt-tuning using frozen/unfrozen LLMs, transfer learning, and few-shot learning abilities.

**Methods**

We developed a soft prompt-based LLM model and compared 4 training strategies including (1) fine-tuning without prompts; (2) hard-prompt with unfrozen LLMs; (3) soft-prompt with unfrozen LLMs; and (4) soft-prompt with frozen LLMs. We evaluated 7 pretrained LLMs using the 4 training strategies for clinical concept and relation extraction on two benchmark datasets. We evaluated the transfer learning ability of the prompt-based learning algorithms in a cross-institution setting. We also assessed the few-shot learning ability.

**Results and Conclusion**

When LLMs are unfrozen, GatorTron-3.9B with soft prompting achieves the best strict F1-scores of 0.9118 and 0.8604 for concept extraction, outperforming the traditional fine-tuning and hard prompt-based models by 0.6~3.1% and 1.2~2.9%, respectively; GatorTron-345M with soft prompting achieves the best F1-scores of 0.8332 and 0.7488 for end-to-end relation extraction, outperforming the other two models by 0.2~2% and 0.6~11.7%, respectively. When LLMs are frozen, small (i.e., 345 million parameters) LLMs have a big gap to be competitive with unfrozen models; scaling LLMs up to billions of parameters makes frozen LLMs competitive with unfrozen

LLMs. For cross-institute evaluation, soft prompting with a frozen GatorTron-8.9B model achieved the best performance. This study demonstrates that (1) machines can learn soft prompts better than humans, (2) frozen LLMs has better few-shot learning ability and transfer learning ability to facilitate muti-institution applications, and (3) frozen LLMs require large models.

**INTRODUCTION**

Pretrained large language models (LLMs) have almost become standard solutions for clinical natural language processing. In the recent decade, the natural language processing (NLP) community has witnessed a dramatic change from fully supervised learning architecture – where the "one model per task" strategy was adopted and all model parameters were tuned (i.e., updated) during training, to pretraining/fine-tuning architecture – where one pretrained LLM can be adapted to various NLP tasks through fine-tuning, and eventually to prompt-based learning architecture [1] – where a prompt was attached to the input to condition the output on not only model parameters but also prompts. Prompt-based learning offers many advantages including better few-shot, zero-shot, and transfer learning ability as well as the freedom to control model output using prompts, which is a key technology in achieving conversational artificial intelligence (AI) such as ChatGPT. [1–3] At present, the performance of prompt-based learning highly depends on (1) the "shape" of the prompts, i.e., hard/discrete prompts (in clear text) or soft/continuous prompts, and (2) algorithms to adopt LLMs for downstream tasks. Currently, there are two strategies to fine-tune LLMs for downstream tasks, including "model-tuning" – updating LLMs parameters in training, or "prompt-tuning" – updating soft prompts while keeping LLMs frozen. Prompt-tuning with frozen LLMs offers many benefits over model-tuning including (1) enabling machines to learn "soft prompts" to unload researchers from labor-intensive prompt engineering (i.e., human manually compose hard prompts using clear text), and (2) reducing computing cost by keeping LLMs frozen; and (3) enabling one model for multiple downstream tasks to greatly reduce the development and deployment cost. Nevertheless, most existing works in prompt-based clinical NLP are based on hard prompts using model-tuning; there is a lack of studies exploring the use of soft prompts and prompt-tuning algorithms. This study seeks to develop a soft prompt-based learning architecture for patient information extraction and to examine the shapes of prompts (i.e., soft prompts, hard prompts) and training strategies (i.e., frozen LLMs, unfrozen LLMs). We systematically examined four different strategies including (1) traditional fine-tuning without prompts, where pretrained LLMs are fine-tuned without prompts; (2) hard prompting with unfrozen LLMs, where human-composed hard prompts are used and LLMs are updated in training; (3) soft prompting with unfrozen LLMs, where parameters from both LLMs and soft prompts are updated in training; and (4) soft prompting with frozen LLMs, where LLMs are frozen and only soft prompts are updated in training. We examined the 4 training strategies for clinical concept extraction and relation extraction using 2 clinical benchmark datasets and 7 clinical LLMs. This study shows that soft prompting can automate the design of prompts through machine learning instead of human engineering and that prompt-tuning using frozen LLMs has better transfer learning and few-shot learning ability than using unfrozen LLMs. Yet, larger LLMs (e.g., models exceed billions of parameters) are required by soft prompting with frozen LLMs. This study provides valuable insights into the selection of prompt shapes and training strategies in adopting LLMs for patient information extraction from clinical narratives.

**BACKGROUND**

Deep learning has profoundly changed clinical NLP in the recent decade. Before deep learning, "full-supervised" learning was widely adopted, where NLP researchers trained traditional machine learning models such as Conditional Random Fields (CRFs) purely using human labeled dataset. Early-stage deep learning models based on convolutional neural networks (CNNs) and recurrent neural networks (RNNs) gradually shifted clinical NLP from fully supervised learning to a pretraining/fine-tuning learning architecture, where models were pretrained using large-scale unlabeled corpora following standard unsupervised learning, and then adapted to various downstream NLP tasks using a fine-tuning step on a small dataset with human labels. In the pretraining/fine-tuning strategy, the parameters of pretrained models are typically updated in the fine-tuning, which is denoted as "model-tuning" in this study. Though researchers can adopt one pretrained model to various downstream tasks through fine-tuning, the fine-tuned models are in fact different from the pretrained model because of model-tuning. Therefore, researchers have to deploy multiple individual models for different NLP tasks in real-world healthcare applications. Later, the transformer-based neural network architecture eventually unified various deep learning models using self-supervised learning with an attention mechanism. The transformer models pretrained using various language models as the learning objectives are known as large language models (LLMs), which is the foundation model in NLP. Along with the success of LLMs, prompt-based learning gradually became mature with promising performances in adopting LLMs to various healthcare applications.

Currently, prompt-based learning is implemented by augmenting the input data with additional information, i.e., prompts, to condition the model output to not only model parameters but also the additional prompts. Prompt-based learning has many advantages with good zero-shot, few-shot, and transfer learning ability, which is a key technology to achieve conversational AI such as ChaptGPT. For example, GPT-3[4] and LAMA[5] directly generate the answers without changing the LLM parameters based on prepended prompts, which are typically composed of a task description and/or several canonical examples. The shape of the prompts and the fine-tuning algorithm are two critical components in adopting prompt-based learning. There are two main types of prompt shapes including (1) "hard prompts" (or discrete prompts) – a piece of text composed by researchers providing information about the target prediction, and (2) "soft prompts" (or continuous prompts) – a continuous vector (of virtual tokens) attached to the input. To adopt LLMs for specific downstream applications, researchers either adopt the traditional fine-tuning to keep updating the pretrained LLMs – known as "model-tuning", or to freeze the LLMs and only update the soft prompts – known as "prompt-tuning" or "p-tuning".[6] Most studies in clinical NLP focus on hard prompts as ChatGPT adopted this strategy and achieved a breakthrough in conversational AI. However, designing hard prompts is very labor-intensive, and recent studies have shown that LLMs are very sensitive to hard prompts. Therefore, many recent studies started exploring prompt-tuning using soft prompts. Prompt-tuning offers many benefits over model-tuning, especially in reducing the computing and memory costs as LLMs are frozen during the fine-tuning. Another important benefit of freezing LLMs in prompt-tuning is that we can deploy one single LLM for multiple tasks in the real-world healthcare applications. However, early-stage studies on prompt-tuning using smaller LLMs have shown that freezing LLMs often does not yield good performance that is competitive with model-tuning. Most recently, several studies further explored prompt-tuning and demonstrated promising results by scaling up the model size to exceed billions of parameters.

LLMs have many potentials in medical research and healthcare. An important application of clinical NLP is patient information extraction from clinical narratives. Clinical concept extraction (or named entity recognition [NER]) and relation extraction (RE) are two fundamental NLP tasks for patient information extraction.[7] Various solutions, including rule-based [8–10] traditional machine learning-based models, [11–14] and deep learning model [15–17] have been developed for clinical concept extraction. The recent transformer architectures, inspired by the self-attention mechanism,[18] have remarkably improved the performance with good ability to manage long-term dependencies and high-level parallelization. Various transformer-based models, such as BERT, ALBERT,[19] RoBERTa, and ELECTRA[20] have been proposed and achieved state-of-the-art performance. Yang *et al.* [21] systematically explored four transformer architectures for clinical concept extraction. Clinical RE is often approached as a classification task to identify relations among clinical concepts. Typically, an end-to-end clinical RE system consists of a first step to identify concepts (i.e., clinical NER) and a second step to classify relations. Both traditional machine learning models [22] [23] [24] and deep learning models, especially transformers, have been explored. We also have explored various transformer models for clinical RE. [25] Though deep learning models, especially transformers, have remarkably improved information extraction [21,26–28], challenges still exists, such as limitations of "BIO" tags in representing overlapped or nested concepts [29] and complexity in enumerating all the candidate concept pairs for relation classification [25]. Recent studies have applied prompt-based learning algorithms to deal with these challenges. Many studies applied a machine reading comprehension (MRC) architecture to train LLMs identify answer spans from the input text according to the information from the prompts.[30–32] For example, Schick et al.[33] introduced Pattern-Exploiting Training (PET) to reformulate input examples into cloze-style phrases. We also developed an MRC model to solve clinical concept extraction and relation extraction in a unified prompt-based model. [30] These existing studies applied hard prompts with model-tuning, where the parameters of LLMs were updated during fine-tuning. However, the design of hard prompts requires intensive validations and domain knowledge. Later, studies explored methods that automatically generate hard prompts using clear text and demonstrated their effectiveness in saving researchers from prompt engineering.[34] Nevertheless, humans cannot enumerate all potential ways to compose the optimal prompts. Liu et al.[35] proposed "prompt-tuning" (or P-tuning) where learnable continuous prompts (i.e., soft prompts) are added to the input embeddings, and parameters from both LLMs and the soft prompts are jointly updated during training. Most recently, more studies have focused on prompt-tuning with frozen LLMs, where the parameters of LLMs are frozen and only the parameters of the soft prompts are updated.[36] For example, Lester et al.[3] proposed adding trainable continuous embeddings to the original input while keeping LLMs unchanged. Liu et al. proposed "P-tuning v2" [6] to add continuous prompts to all layers of the LLM and demonstrated that prompt-tuning can be comparable to fine-tuning across a wide range of model scales and NLP tasks.

Inspired by recent studies exploring soft prompts with frozen LLMs, this study seeks to systematically examine the shapes of prompts (i.e., hard or soft) and fine-tuning strategies (i.e., frozen or unfrozen LLMs) in clinical concept extraction and relation extraction. We systematically explored 4 different strategies involving different shapes of prompts and fine-tuning algorithms. We developed a soft-prompt based MRC architecture and explored 7 pretrained LLMs with different sizes to examine the 4 different strategies on two benchmark datasets for clinical concepts and relation extraction.

## MATERIALS AND METHODS

**Dataset**

This study used two clinical benchmark datasets for clinical concept extraction and relation extraction, including the 2018 n2c2 dataset (track 2) focusing on the extraction of adverse drug events and medication (referred to as the drug-ADE dataset) [37], and the 2022 n2c2 dataset (track 2) focusing on the extraction of social determinant of health (referred to as the SDoH dataset) [38]. The drug-ADE dataset consists of 505 discharge summaries taken from the Medical Information Mart for Intensive Care (MIMIC)-III database with annotations of 9 categories of clinical concepts (drug, drug attributes, ADEs) and 8 categories of relations among drugs, drug-associated attributes, and ADEs. The SDoH dataset consists of 5 categories of SDoH concepts and 9 categories of SDoH-associated attribute concepts, and 28 categories of relations among SDoH concepts and SDoH-associated attributes. The 2022 n2c2 challenge provides two datasets including the MIMIC dataset (the clinical notes are taken from the MIMIC-III database), and the University of Washington dataset (the clinical notes are taken from UW). Table 1 shows summary statistics of the two datasets.

**Table 1.** Summary statistics of the clinical notes and annotated concepts and relations in 2018 n2c2 drug-ADE dataset and the 2022 n2c2 SDoH datasets.

| Challenge | Datasets | Number of notes | Number of clinical concepts | Number of clinical relations |
|---|---|---|---|---|
| 2018 n2c2 (Medication-ADE) | Training | 303 | 50951 | 36384 |
| | Test | 202 | 32918 | 23462 |
| 2022 n2c2 (SDoH, attributes) | Training (MIMIC-III) | 1316 | 16039 | 10933 |
| | Development (MIMIC-III) | 188 | 1744 | 1177 |
| | Test (MIMIC-III) | 373 | 3331 | 2243 |
| | UW-test | 518 | 4903 | 3249 |

ADE: adverse drug event; SDoH: social determinants of health: UW: University of Washington; MIMIC-III: Medical Information Mart for Intensive Care.

**Strategies for prompt shapes and fine-tuning algorithms**

Based on a hard prompt-based MRC model developed in our previous study[30], we replaced the hard prompts into soft prompts and developed prompt-tuning algorithms. Specifically, we added a continuous and learnable task-specific soft prompt to instruct LLMs to identify both concepts and relations. We investigated four different strategies including (1) traditional pretraining/fine-tuning without prompts; (2) model-tuning using hard prompts; (3) prompt-tuning using soft prompts, where both LLMs and prompts were updated; and (4) prompt-tuning using soft prompts with LLMs freezing. As previous studies showed that the size of LLMs is very important to prompt-tuning, we explored 7 pre-trained LLMs with different sizes up to 3.9 billion and 8.9 billion parameters. We also examined the few-shot learning ability and the transfer learning ability of the four strategies. Figure 1 shows an overview of the four strategies.

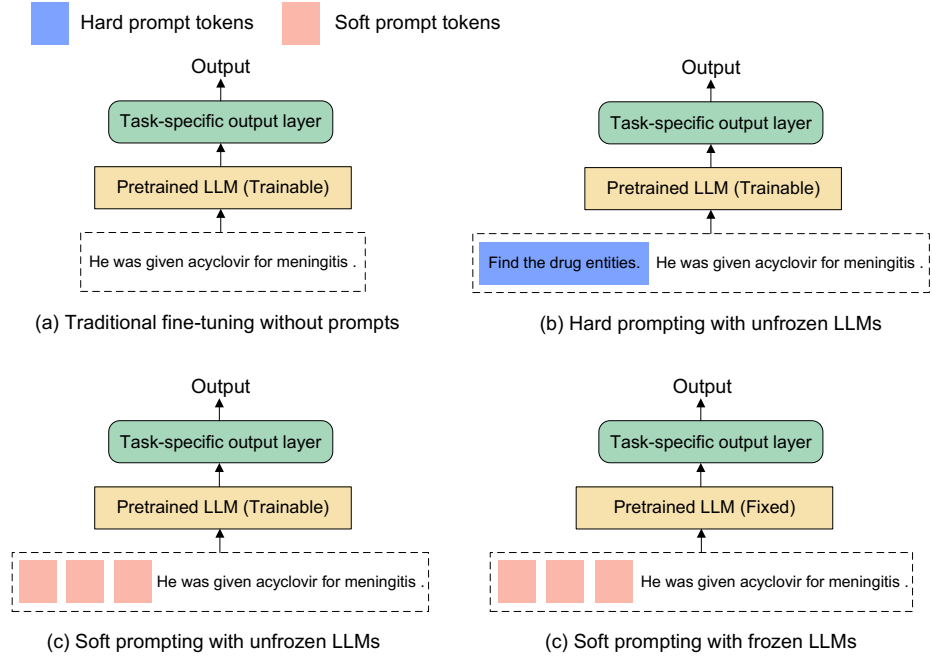

**Figure 1**. Comparison of different pre-trained transformer model tuning paradigms. (a) Traditional fine-tuning without prompts; (b) hard prompting with unfrozen LLMs; (c) soft prompting with unfrozen LLMs; (d) soft prompting with frozen LLMs.

**Soft prompt-based MRC for clinical concept and relation extraction**

In a previous study[30], we have developed a MRC model using hard prompts (i.e., questions) such as "What drug is mentioned in the text?" to instruct LLMs identify drug concepts. Here we extend this MRC model to use soft prompts. We adopt trainable soft prompts as a substitution for hand-designed hard prompts. Given the input text $X = \{x_1, x_2, \ldots, x_n\}$, a series of prompt tokens $P = \{p_1, p_2, \ldots, p_m\}$ is prepended to the right of input text, then the input embedding sequence can be written as $[h_1, h_2, \ldots, h_m, e(x_1), e(x_2), \ldots, e(x_n)]$, where $h_i$ is the trainable embedding tensors mapped from $p_i$, and $x_i$ is mapped to $e(x_i)$ by the pre-trained embedding layer. To achieve more fine-grained control over LLMs, we adopted the idea of deep prompt turning,[6] where the trainable prompts are added to every layers in addition to the input layer.

As shown in Figure 2, the MRC model first identified a trigger concept $e_i$ (e.g., "Drug" in the drug-ADE dataset), where the prompts of the trigger concept are prepended to every layers. Then we leveraged the extracted trigger concepts to identify the attribute concepts and the relations in a similar manner but introduced a verbalizer to define the trigger concept spans using two anchor tokens (i.e., [S] and [E]). The final hidden representations of each token are used for the entity span prediction. Similar to the previous MRC model, we introduced two binary classifiers, one to predict whether each token is the start index or not, and the other to predict whether each token is the end index. We used a third classifier to match the start index to the corresponding end index when there are multiple answers. The detailed model architecture and loss functions are provided in the supplementary material.

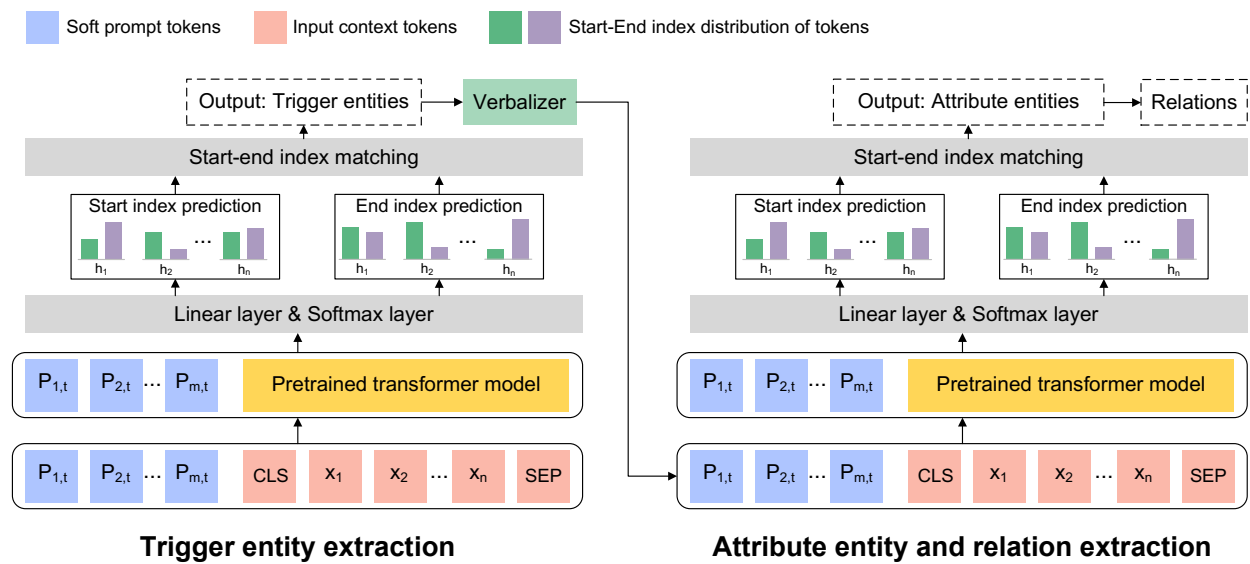

Figure 2. Overall architecture of soft prompt-based transformer architecture for clinical concept extraction and end-to-end relation extraction. The verbalizer is introduced to add anchor tokens to the extracted trigger entities to identify relations between the trigger concept and other concepts.

**Transfer learning and few-shot learning**

We examined the transfer learning ability by comparing the performance in 2 evaluation settings including (1) unified institution: training on MIMIC-train and testing on MIMIC-test, (2) cross-institution: training on MIMIC-train and testing on UW-test. We also compared the 3 prompt-based learning algorithms to evaluate their few-shot learning ability using the 2022 n2c2 SDoH dataset. Following standard setting for few-shot learning, we evaluated the 3 prompt-based learning algorithms using 5, 10, 20, 50, and 100 training samples for each category of SDoH.

**Experiments and evaluation**

We explored 7 LLMs including BERT,[39] BERT-MIMIC,[21] RoBERTa,[40] RoBERTa-MIMIC,[21] GatorTron-base of 345 million parameters (denoted as GatorTron-345M), and GatorTron-medium of 3.9 billion parameters (denoted as GatorTron-3.9B), and GatorTron-large of 8.9 billion parameters (denoted as GatorTron-8.9B).[41]. Our previous studies showed that BERT, BERT-MIMIC, RoBERTa and RoBERTa-MIMIC are among the top-performing transformer models for clinical concept and relation extraction. GatorTron is currently the largest clinical LLM pre-trained using over 90 billion words of text and achieved state-of-the-art performance for five clinical NLP tasks.[41] We used the official evaluation scripts provided by the 2018 n2c2 challenges to evaluate the strict F1-score for standard clinical concept and relation extraction. We examined the generalizability of the prompt learning models when applied to cross-institution settings using the 2022 n2c2 MIMIC-train and the UW-test datasets. We explored the few-shot learning ability of the prompting learning models using the 2022 n2c2 dataset. We also conducted ablation studies to explore the impact of prompt length on the model performance using the 2022 n2c2 dataset.

BERT: a bidirectional transformer-based encoder language model pre-trained over a large general English domain corpus. BERT adopted masked language modeling (MLM) and next-sentence prediction (NSP) training objectives to create deep representations capturing contextual information.

RoBERTa: a transformer-based model with the same architecture as BERT but pre-trained using dynamic masked language modeling and optimized using different strategies.

GatorTron: a large clinical language model developed in our previous work, which is pre-trained from scratch using >90 billion words of text (including >82 billion words of de-identified clinical text). In our experiment, we used the three GatorTron models which contain 345 million, 3.9 billion and 8.9 billion parameters, respectively.

***Experiment settings:*** We developed the soft prompt-based MRC model using the Transformers library developed by the HuggingFace team using the PyTorch Lightning library. For pretrained models from the general English domain, we used the default models hosted at the HuggingFace model repository. For clinical transformer models, we used the existing models in our previous study,[21] including BERT-MIMIC and RoBERTa-MIMIC pre-trained using the MIMIC-III corpus, and GatorTron pre-trained using UF Health clinical notes. We adopted a five-fold cross-validation strategy to optimize hyperparameters, including the learning rate, the training batch size, and the training loss weight. The best models were selected according to the cross-validation performances measured by micro-averaged strict F1-score. All experiments were conducted using 8 Nvidia A100-80G GPUs.

# RESULTS

Table 2 compares 7 LLMs (strict micro-averaged F1-score) using 4 training strategies for clinical concept extraction on two benchmark datasets developed by the 2018 n2c2 and the 2022 n2c2 challenges. For the 2018 n2c2 drug-ADE dataset, soft-prompt unfrozen GatorTron-3.9B model achieved the best F1-score of 0.9118, outperforming traditional pretraining/fine-tuning and hard prompt-based MRC models with an average improvement of 2.4% and 0.6%, respectively. When LLMs are frozen, the models using smaller LLMs with 345 million parameters or under dropped by approximately 5% compared with the corresponding models with unfrozen LLMs; whereas the GatorTron-3.9B and GatorTron-8.9B models achieved comparable F1-scores of 0.9085 and 0.9093, respectively. We observed similar results for the 2022 n2c2 SDoH dataset, where the soft prompt GatorTron-8.9B model achieved the best F1-score of 0.8610, outperforming the traditional pretraining/fine-tuning models and hard prompt-based MRC models by 0.9~2.3%. When LLMs are frozen, smaller LLMs with 345 million parameters or under remarkably dropped 3.8% ~ 5.3%, whereas larger LLMs, GatorTron-3.9B and GatorTron-8.9B, achieved comparable F1-scores of 0.8579 and 0.8588, respectively.

**Table 2.** Comparison of 7 LLMs using 4 fine-tuning strategies for clinical concept extraction.

| Corpus | Models | Training strategies | | | |
|---|---|---|---|---|---|
| | | No prompt; Unfrozen LLM | Hard prompt; Unfrozen LLM | Soft prompt; Unfrozen LLM | Soft prompt; Frozen LLM |
| 2018 n2c2 | BERT | 0.8807 | 0.9018 | 0.9095 | 0.8598 |
| | BERT-MIMIC | 0.8853 | 0.9031 | 0.9080 | 0.8676 |
| | RoBERTa | 0.8812 | 0.9016 | 0.9076 | 0.8490 |
| | RoBERTa-MIMIC | **0.8907** | 0.9020 | 0.9081 | 0.8621 |
| | GatorTron-345M | 0.8879 | 0.9059 | 0.9112 | 0.8659 |
| | GatorTron-3.9B | 0.8883 | 0.9051 | **0.9118** | 0.9085 |
| | GatorTron-8.9B | **0.8891** | **0.9063** | 0.9115 | **0.9093** |
| 2022 n2c2 | BERT | 0.8318 | 0.8424 | 0.8549 | 0.8019 |
| | BERT-MIMIC | 0.8378 | 0.8435 | 0.8568 | 0.8128 |
| | RoBERTa | 0.8331 | 0.8424 | 0.8523 | 0.8102 |
| | RoBERTa-MIMIC | 0.8339 | 0.8412 | 0.8537 | 0.8129 |
| | GatorTron-345M | 0.8341 | 0.8451 | 0.8571 | 0.8186 |
| | GatorTron-3.9B | 0.8370 | **0.8482** | 0.8604 | 0.8579 |
| | GatorTron-8.9B | **0.8388** | 0.8478 | **0.8610** | **0.8588** |

No prompt: No prompt was added to the input; Hard prompt: discrete prompts designed by researchers as clear text; Soft prompts: a trainable continuous vector; Model-tuning: the parameters of LLMs are updated in fine-tuning; Prompt-tuning: the trainable continuous vectors of soft prompts are updated in the fine-tuning.

**Table 3**. Comparison of 7 LLMs using 4 fine-tuning strategies for end-to-end clinical relation extraction.

| Corpus | Models | Training strategies | | | |
|---|---|---|---|---|---|
| | | No prompt, Unfrozen LLM | Hard prompt, Unfrozen LLM | Soft prompt, Unfrozen LLM | Soft prompt, Frozen LLM |
| 2018 n2c2 | BERT | 0.8129 | 0.8251 | 0.8238 | 0.7820 |
| | BERT-MIMIC | 0.8141 | 0.8279 | 0.8301 | 0.7931 |
| | RoBERTa | 0.8132 | 0.8258 | 0.8272 | 0.7887 |
| | RoBERTa-MIMIC | 0.8150 | 0.8261 | 0.8297 | 0.7903 |
| | GatorTron-345M | 0.8192 | 0.8291 | **0.8332** | 0.7921 |
| | GatorTron-3.9B | **0.8205** | 0.8283 | 0.8321 | **0.8299** |
| | GatorTron-8.9B | 0.8200 | **0.8310** | 0.8330 | 0.8268 |
| 2022 n2c2 | BERT | 0.6322 | 0.7325 | 0.7386 | 0.6852 |
| | BERT-MIMIC | 0.6401 | 0.7426 | 0.7481 | 0.7011 |
| | GatorTron-345M | 0.6395 | 0.7364 | **0.7488** | 0.7035 |
| | GatorTron-3.9B | 0.6524 | 0.7356 | 0.7450 | **0.7442** |
| | GatorTron-8.9B | **0.6601** | **0.7430** | 0.7461 | 0.7432 |

No prompt: No prompt was added to the input; Hard prompt: discrete prompts designed by researchers as clear text; Soft prompts: a trainable continuous vector; Model-tuning: the parameters of LLMs are updated in fine-tuning; Prompt-tuning: the trainable continuous vectors of soft prompts are updated in the fine-tuning.

Table 3 compares 7 LLMs (strict micro-averaged F1-score) using 4 training strategies for end-to-end clinical relation extraction on the two benchmark datasets. For the drug-ADE relation extraction, all soft prompt unfrozen LLMs outperformed traditional fine-tuning and hard prompt-based LLMs with 0.2 ~ 1.7%. The GatorTron-345M model achieved the best F1-score of 0.8332. When LLMs were frozen, the performance of smaller LLMs (with 345 million parameters or under) remarkably dropped by 3.7~4.7%, including the GatorTron-345M model; whereas the GatorTron-3.9B and GatorTron-8.9B models achieved comparable F1-scores of 0.8299 and 0.8268, respectively. We observed large gaps from the SDoH dataset, where the soft prompt LLMs outperformed traditional fine-tuning by 8.6~10.6% and outperformed the hard prompt-based MRC models by 0.3~1.2%, respectively. GatorTron-345M model achieved the best F1-score of 0.7488. When LLMs were frozen, LLMs with 345 million or under remarkably dropped by 4.1~5.9%; whereas the GatorTron-3.9B and GatorTron-8.9B models achieved F1-scores of 0.7442 and 0.7432, comparable to unfrozen LLMs.

Table 4 compares GatorTron-3.9B and GatorTron-8.9B for cross-institution evaluation. When the training and test datasets are both from MIMIC, soft prompting with unfrozen LLMs achieved the best performance. Nevertheless, when the test data is from a different institution (i.e., UW-test), soft prompting with frozen LLMs achieved the best performance; GatorTron-3.9B outperformed the other two models by 1.3~1.7% and GatorTron-8.9B outperformed the other two models by 1.2~1.6%, demonstrating better transfer learning ability of using frozen LLMs for cross-institute applications.

**Table 4**. Cross-institute comparison of GatorTron-3.9B and GatorTron-8.9B models trained using different prompt-based learning strategies on the 2022 n2c2 SDoH dataset.

| Model | Training | Test | Training strategy | SDoH concepts and attributes extraction | End-to-end relation extraction |
|---|---|---|---|---|---|
| GatorTron-3.9B | MIMIC-train | MIMIC-test | Hard prompt, Unfrozen LLM | 0.8482 | 0.7364 |
| | | | Soft prompt, Unfrozen LLM | **0.8604** | **0.7450** |
| | | | Soft prompt, Frozen LLM | 0.8579 | 0.7442 |
| | MIMIC-train | UW-test | Hard prompt, Unfrozen LLM | 0.8124 | 0.7122 |
| | | | Soft prompt, Unfrozen LLM | 0.8129 | 0.7139 |
| | | | Soft prompt, Frozen LLM | **0.8297** | **0.7266** |
| GatorTron-8.9B | MIMIC-train | MIMIC-test | Hard prompt, Unfrozen LLM | 0.8478 | 0.7430 |
| | | | Soft prompt, Unfrozen LLM | **0.8610** | **0.7461** |
| | | | Soft prompt, Frozen LLM | 0.8588 | 0.7432 |
| | MIMIC-train | UW-test | Hard prompt, Unfrozen LLM | 0.8139 | 0.7141 |
| | | | Soft prompt, Unfrozen LLM | 0.8145 | 0.7162 |
| | | | Soft prompt, Frozen LLM | **0.8299** | **0.7280** |

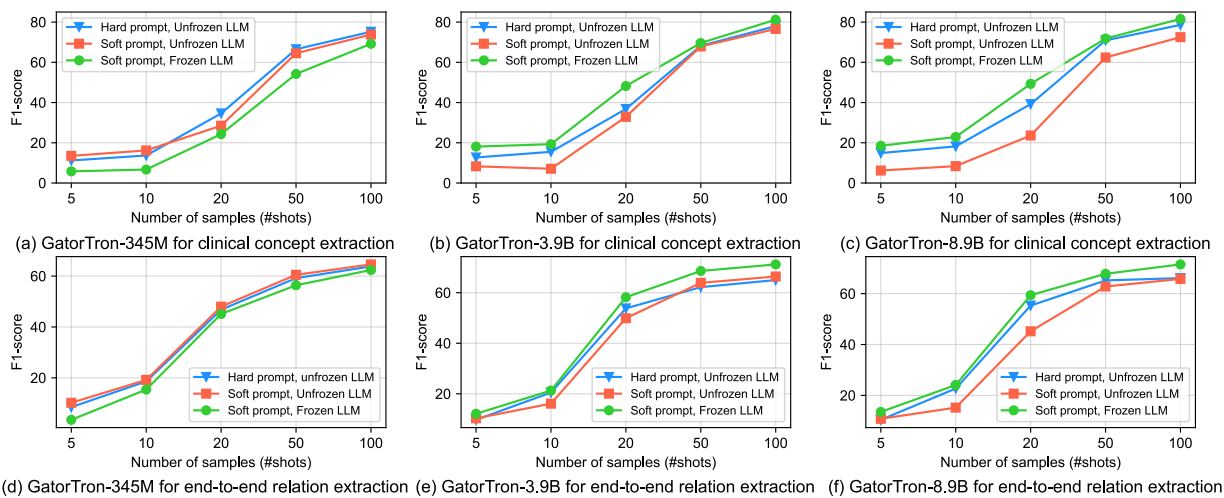

**Figure 3**. Comparison of few-shot learning performance of different prompting algorithms for SDoH concept and relation extraction (MIMIC-train, MIMIC-test).

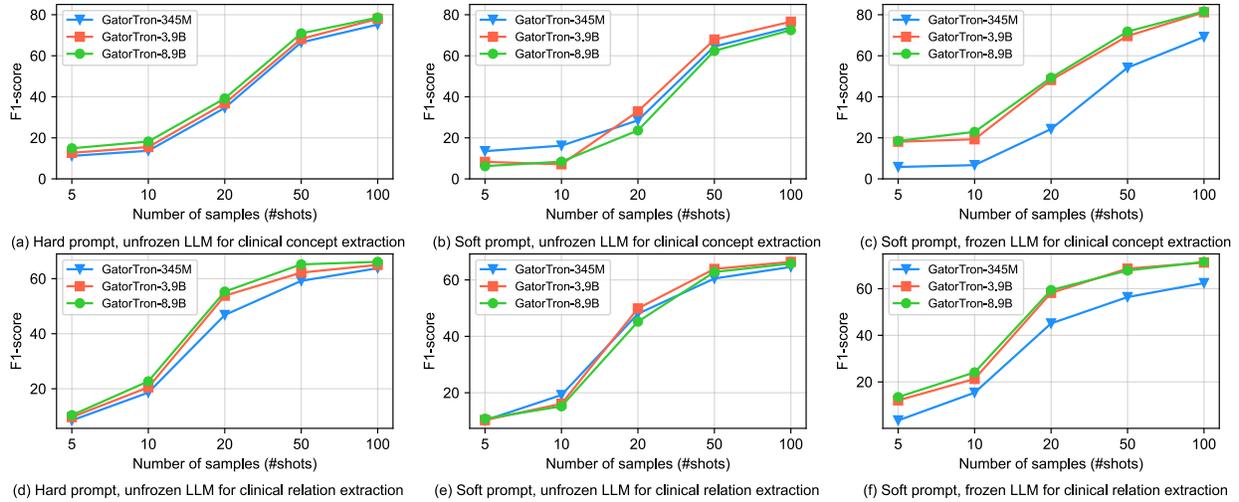

**Figure 4.** Comparison of few-shot learning performance of GatorTron models for SDoH concept and relation extraction (MIMIC-train, MIMIC-test).

Figure 3 compares few-shot learning of different prompting algorithms for SDoH concept and relation extraction. When using a smaller GatorTron-345M model, prompting (both hard and soft prompts) with unfrozen LLMs has better few-short learning ability than prompting with frozen LLMs. However, when using larger GatorTron models over 3.9 billion parameters, soft prompting with frozen LLMs has better few-shot learning performance. Figure 4 compares the few-shot learning performance of different GatorTron models. When LLMs were unfrozen, the 3 GatorTron models have no big difference in few-shot learning performance. However, when LLMs are frozen, larger models, i.e., GatorTron-3.9B and GatorTron-8.9B, demonstrated better few-shot learning performance than the smaller model. By using 100 training samples, soft prompting with frozen GatorTron-8.9B achieved F1 scores of 81.6% and 71.5% for two tasks, respectively, which are comparable to the best F1 scores of 85.88% and 74.32% achieved by training with the full-dataset.

# DISCUSSION

Prompt-based learning is a key technology to adopt LLMs for various downstream clinical NLP applications. The shape of prompts (e.g., hard or soft) and the training strategies (e.g., frozen LLM or unfrozen LLM) are critical elements for a robust prompt-based learning system. Most existing studies mainly focus on hard prompts and play with existing general-purpose LLMs (e.g., ChatGPT) by feeding hard prompts, i.e., prompt engineering. There is a lack of studies to examine the shapes of prompts and prompt-tuning methods to unload researchers from time-consuming prompt engineering. This study developed a soft prompt-based MRC model, systematically examined 4 different strategies to adopt LLMs for clinical concept extraction and clinical relation extraction and compared 7 LLMs of various sizes using two benchmark datasets. We also evaluated the transfer learning ability and few-shot learning ability. The experimental results show that the proposed soft prompt-based MRC model achieved state-of-the-art performance for extracting relations of drugs and ADEs as well as SDoH, outperforming traditional fine-tuning models and hard prompt-based models. This study shows that machines could learn better "soft prompts" that outperform hard prompts composed by humans, demonstrating the efficiency of using soft prompts to adopt LLMs for various clinical applications.

Soft prompting with frozen LLMs is very attractive as it enables the deployment of a single LLM to solve multiple applications. Most existing studies focus on hard prompts and prompt-based learning using unfrozen LLMs, which eventually generate multiple task-specific LLMs for various applications as the LLMs are updated during the training. This unfrozen strategy not only requires huge computing cost to update LLMs but also cause a substantial burden to deploy multiple task-specific LLMs in real-world clinical applications. Training with frozen LLMs is more parameter-efficient since only a small set of parameters of the prompts (e.g., 2.5~6% in our experiments) are updated during the training, which greatly reduces the computing cost. Yet, to enjoy this benefit of frozen LLMs, large LLMs with parameters over billions of parameters are required, such as the GatorTron-3.9B and GatorTron-8.9B. For small LLMs (e.g., 345 million parameters), prompting with frozen LLMs has a big gap to be competitive with prompting with unfrozen LLMs. Scaling up the size of LLMs could narrow this gap and eventually achieve comparable performance with prompting with unfrozen LLMs for clinical concept extraction and relation extraction. One possible reason is that larger LLMs possess a more expansive parameter space to better capture and reflect information from soft prompts, thereby providing better resolution in prediction. This finding is very important to instruct future deployment of LLMs in real-world clinical applications. As more and more AI modules are expected to be integrated into the EHR systems, there will be a serious deployment issue in using traditional pre-training/fine-tuning strategies as multiple models are required for different tasks, which will cause huge deployment burden as well as maintenance cost for healthcare IT. Prompting with frozen LLMs will enable the deployment of one LLM for multiple applications to prevent such deployment burden and maintenance costs.

This study also shows that soft prompting using frozen LLMs has better transfer learning and few-shot learning ability to improve cross-institution applications and reduce annotation costs. When the training and test datasets are from the same MIMIC corpus, prompting using unfrozen LLMs performs slightly better than prompting with frozen LLMs; however, when the training and test datasets are from two different institutions, prompting with frozen LLMs is remarkably better than prompting with unfrozen LLMs. When the model size increases from 3.9 billion to 8.9 billion, the performance of both tasks improves, indicating that larger LLMs have better transfer learning ability for cross-institution applications. One potential reason is that prompting with unfrozen LLMs will potentially overfit the pretrained model to the training data of a specific institution, whereas prompting with frozen LLMs could keep the generalizability of LLMs to better support cross-institution applications. Our study also demonstrates that soft prompting with frozen LLMs has better few-shot learning performance outperforming traditional pre-training/finetuning and prompting with unfrozen LLMs when the LLMs have over billions of parameters. Our results show that the few-shot learning and the transfer learning are associated with the size of the model as larger LLMs have more parameters to better deal with unseen samples.

We conducted an error analysis to compare soft prompting with hard prompting and pretraining/fine-tuning and discovered that soft prompting performs better for overlapped or nested concepts and their relations. For example, the soft prompting using unfrozen GatorTron-345M identified 39% and 82% of the overlapped/nested concepts, outperforming the corresponding GatorTron-345M model trained using traditional pre-training/fine-tuning, which only accurately identified 28% and 9%, respectively. We also examined how the length of the soft prompts affect the performance. Figure 5 compares the performance of GatorTron-3.9B model using soft prompts with different lengths (8, 16, 32, 64, and 128) using the 2022 n2c2 SDoH dataset. We can see that soft prompting is sensitive to the length of prompts, which could cause performance differences ranging from 1% to 2%. Overall, moderate-length prompts (i.e., 32 and 64) achieved better F1 scores compared with extremely short or long prompts. In addition, the length of the soft prompts has a more remarkable impact when prompting with frozen LLMs as only the parameters from prompts were updated.

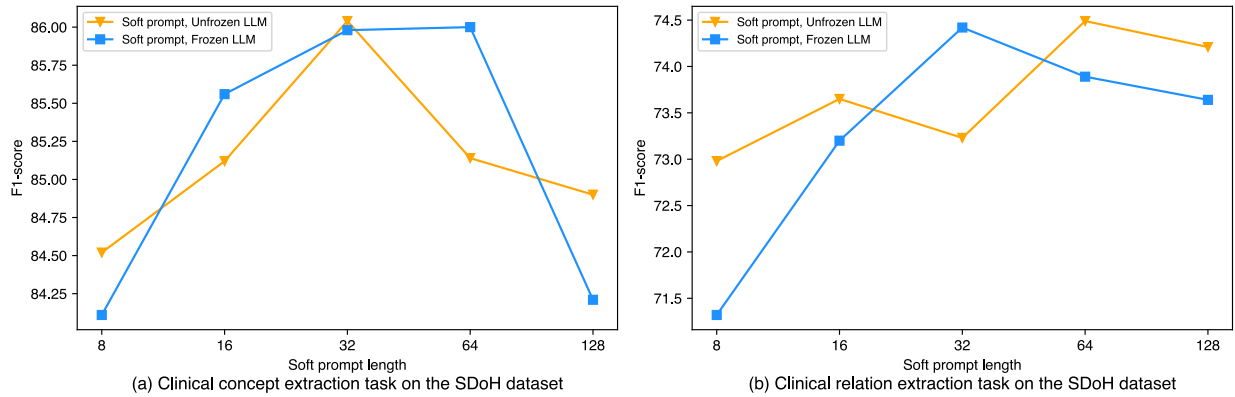

**Figure 5**. Impact of different soft prompt lengths for GatorTron-3.9B model with different model tuning strategies.

This study has limitations. We mainly focus on encoder-only LLMs based on BERT architecture, future studies should examine the generative LLMs based on decoder architectures such as GPT-3 model and encoder-decoder models such as T-5 [42]. In addition, it's of great value to explore parameter-efficient model tuning approaches as fine-tuning LLMs is computationally expensive and larger LLMs may be overkill for some applications. Our future work will investigate decoder-only LLMs and encoder-decoder LLMs.

## CONCLUSION

This study investigated model-tuning and prompt-tuning strategies to adopt LLMs for clinical concept extraction and relation extraction. The proposed soft prompting with frozen LLMs is parameter efficient to enable machines to learn soft prompts to unload researchers from labor-intensive prompt engineering. Prompting with frozen LLMs also has better transfer learning and few-shot learning ability to better facilitate cross-institution applications and reduce annotation costs.

ACKNOWLEDGMENTS

We would like to thank the i2b2 and n2c2 challenge organizers to provide the annotated corpus. We gratefully acknowledge the support of NVIDIA Corporation and the NIVIDA AI Technology Center (NVAITC) UF program with the donation of the GPUs used for this research.

FUNDING STATEMENT

This study was partially supported by a Patient-Centered Outcomes Research Institute® (PCORI®) Award (ME-2018C3-14754), a grant from the National Cancer Institute, 1R01CA246418 R01, a grant from the National Institute on Aging, NIA R21AG062884, and the Cancer Informatics and eHealth core jointly supported by the UF Health Cancer Center and the UF Clinical and Translational Science Institute. The content is solely the responsibility of the authors and does not necessarily represent the official views of the funding institutions.

COMPETING INTERESTS STATEMENT

Cheng Peng, Xi Yang, Kaleb E Smith, Zehao Yu, Aokun Chen, Jiang Bian, and Yonghui Wu have no conflicts of interest that are directly relevant to the content of this study.

CONTRIBUTORSHIP STATEMENT

CP, XY and YW were responsible for the overall design, development, and evaluation of this study. CP, ZY and KS performed the experiments. CP and YW did the initial drafts of the manuscript, XY, AC, and JB also contributed to writing and editing of this manuscript. All authors reviewed the manuscript critically for scientific content, and all authors gave final approval of the manuscript for publication.

## SUPPLEMENTARY MATERIAL

Attached in a separate document.

## DATA AVAILABILITY

The drug-ADE and SDoH data sets used in this work were provided by the n2c2 organizers (see https://n2c2.dbmi.hms.harvard.edu/ for details).

## REFERENCES

1  Liu P, Yuan W, Fu J, *et al.* Pre-train, prompt, and predict: A systematic survey of prompting methods in natural language processing. *ACM Comput Surv* 2023;**55**:1–35.

2  Liu X, Ji K, Fu Y, *et al.* P-tuning: Prompt tuning can be comparable to fine-tuning across scales and tasks. In: *Proceedings of the 60th Annual Meeting of the Association for Computational Linguistics (Volume 2: Short Papers)*. Stroudsburg, PA, USA: : Association for Computational Linguistics 2022. doi:10.18653/v1/2022.acl-short.8

3  Lester B, Al-Rfou R, Constant N. The power of scale for parameter-efficient prompt tuning. In: *Proceedings of the 2021 Conference on Empirical Methods in Natural Language Processing*. Stroudsburg, PA, USA: : Association for Computational Linguistics 2021. doi:10.18653/v1/2021.emnlp-main.243

4  Brown TB, Mann B, Ryder N, *et al.* Language Models are Few-Shot Learners. arXiv [cs.CL]. 2020.http://arxiv.org/abs/2005.14165

5  Petroni F, Rocktäschel T, Riedel S, *et al.* Language models as knowledge bases? In: *Proceedings of the 2019 Conference on Empirical Methods in Natural Language Processing and the 9th International Joint Conference on Natural Language Processing (EMNLP-IJCNLP)*. Stroudsburg, PA, USA: : Association for Computational Linguistics 2019. doi:10.18653/v1/d19-1250

6	Liu X, Ji K, Fu Y, *et al.* P-tuning v2: Prompt tuning can be comparable to fine-tuning universally across scales and tasks. arXiv [cs.CL]. 2021.http://arxiv.org/abs/2110.07602

7	Fraile Navarro D, Ijaz K, Rezazadegan D, *et al.* Clinical named entity recognition and relation extraction using natural language processing of medical free text: A systematic review. *Int J Med Inform* 2023;**177**:105122.

8	Esuli A, Marcheggiani D, Sebastiani F. An enhanced CRFs-based system for information extraction from radiology reports. *J Biomed Inform* 2013;**46**:425–35.

9	Childs LC, Enelow R, Simonsen L, *et al.* Description of a rule-based system for the i2b2 challenge in natural language processing for clinical data. *J Am Med Inform Assoc* 2009;**16**:571–5.

10	Wyles CC, Tibbo ME, Fu S, *et al.* Use of natural language processing algorithms to identify common data elements in operative notes for total hip arthroplasty. *J Bone Joint Surg Am* 2019;**101**:1931–8.

11	Clark C, Good K, Jezierny L, *et al.* Identifying smokers with a medical extraction system. *J Am Med Inform Assoc* 2008;**15**:36–9.

12	Doan S, Xu H. Recognizing Medication related Entities in Hospital Discharge Summaries using Support Vector Machine. *Proc Int Conf Comput LING* 2010;**2010**:259–66.

13	Gaebel J, Kolter T, Arlt F, *et al.* Extraction of adverse events from clinical documents to support decision making using semantic preprocessing. *Stud Health Technol Inform* 2015;**216**:1030.

14	Forsyth AW, Barzilay R, Hughes KS, *et al.* Machine learning methods to extract documentation of breast cancer symptoms from electronic health records. *J Pain Symptom Manage* 2018;**55**:1492–9.

15	Tan LK, Liew YM, Lim E, *et al.* Convolutional neural network regression for short-axis left ventricle segmentation in cardiac cine MR sequences. *Med Image Anal* 2017;**39**:78–86.

16	Jauregi Unanue I, Zare Borzeshi E, Piccardi M. Recurrent neural networks with specialized word embeddings for health-domain named-entity recognition. *J Biomed Inform* 2017;**76**:102–9.

17	Baytas IM, Xiao C, Zhang X, *et al.* Patient subtyping via time-aware LSTM networks. In: *Proceedings of the 23rd ACM SIGKDD International Conference on Knowledge Discovery and Data Mining*. New York, NY, USA: : ACM 2017. doi:10.1145/3097983.3097997

18	Vaswani A, Shazeer N, Parmar N, *et al.* Attention is all you need. *Adv Neural Inf Process Syst* 2017;**30**.https://proceedings.neurips.cc/paper/7181-attention-is-all

19	Lan Z, Chen M, Goodman S, *et al.* ALBERT: A lite BERT for self-supervised learning of language representations. arXiv [cs.CL]. 2019.http://arxiv.org/abs/1909.11942

20  Clark K, Luong M-T, Le QV, *et al.* ELECTRA: Pre-training text encoders as discriminators rather than generators. arXiv [cs.CL]. 2020.http://arxiv.org/abs/2003.10555

21  Yang X, Bian J, Hogan WR, *et al.* Clinical concept extraction using transformers. *J Am Med Inform Assoc* 2020;**27**:1935–42.

22  Tang B, Cao H, Wu Y, *et al.* Clinical entity recognition using structural support vector machines with rich features. In: *Proceedings of the ACM sixth international workshop on Data and text mining in biomedical informatics*. New York, NY, USA: : ACM 2012. doi:10.1145/2390068.2390073

23  Chapman AB, Peterson KS, Alba PR, *et al.* Detecting adverse drug events with rapidly trained classification models. *Drug Saf* 2019;**42**:147–56.

24  Bose P, Srinivasan S, Sleeman WC IV, *et al.* A survey on recent named Entity Recognition and Relationship Extraction techniques on clinical texts. *Appl Sci (Basel)* 2021;**11**:8319.

25  Yang X, Yu Z, Guo Y, *et al.* Clinical relation extraction using transformer-based models. arXiv [cs.CL]. 2021.http://arxiv.org/abs/2107.08957

26  Fu S, Chen D, He H, *et al.* Clinical concept extraction: A methodology review. *J Biomed Inform* 2020;**109**:103526.

27  Li Y, Wehbe RM, Ahmad FS, *et al.* Clinical-Longformer and Clinical-BigBird: Transformers for long clinical sequences. arXiv [cs.CL]. 2022.http://arxiv.org/abs/2201.11838

28  Suárez-Paniagua V, Rivera Zavala RM, Segura-Bedmar I, *et al.* A two-stage deep learning approach for extracting entities and relationships from medical texts. *J Biomed Inform* 2019;**99**:103285.

29  Ju M, Nguyen NTH, Miwa M, *et al.* An ensemble of neural models for nested adverse drug events and medication extraction with subwords. *J Am Med Inform Assoc* 2020;**27**:22–30.

30  Peng C, Yang X, Yu Z, *et al.* Clinical concept and relation extraction using prompt-based machine reading comprehension. *J Am Med Inform Assoc* Published Online First: 14 June 2023. doi:10.1093/jamia/ocad107

31  Li X, Feng J, Meng Y, *et al.* A unified MRC framework for named entity recognition. In: *Proceedings of the 58th Annual Meeting of the Association for Computational Linguistics*. Stroudsburg, PA, USA: : Association for Computational Linguistics 2020. doi:10.18653/v1/2020.acl-main.519

32  Li X, Yin F, Sun Z, *et al.* Entity-relation extraction as multi-turn question answering. arXiv [cs.CL]. 2019.http://arxiv.org/abs/1905.05529

33  Schick T, Schütze H. Few-shot text generation with pattern-exploiting training. arXiv [cs.CL]. 2020.http://arxiv.org/abs/2012.11926

34  Jiang Z, Xu FF, Araki J, *et al.* How can we know what language models know? *Trans Assoc Comput Linguist* 2020;**8**:423–38.

35  Liu X, Zheng Y, Du Z, *et al.* GPT Understands, Too. arXiv [cs.CL]. 2021.http://arxiv.org/abs/2103.10385

36  Li XL, Liang P. Prefix-tuning: Optimizing continuous prompts for generation. In: *Proceedings of the 59th Annual Meeting of the Association for Computational Linguistics and the 11th International Joint Conference on Natural Language Processing (Volume 1: Long Papers)*. Stroudsburg, PA, USA: : Association for Computational Linguistics 2021. doi:10.18653/v1/2021.acl-long.353

37  Henry S, Buchan K, Filannino M, *et al.* 2018 n2c2 shared task on adverse drug events and medication extraction in electronic health records. *J Am Med Inform Assoc* 2020;**27**:3–12.

38  Lybarger K, Yetisgen M, Uzuner Ö. The 2022 n2c2/UW shared task on extracting social determinants of health. *J Am Med Inform Assoc* Published Online First: 16 February 2023. doi:10.1093/jamia/ocad012

39  Devlin J, Chang M-W, Lee K, *et al.* BERT: Pre-training of deep bidirectional Transformers for language understanding. arXiv [cs.CL]. 2018.http://arxiv.org/abs/1810.04805

40  Liu Y, Ott M, Goyal N, *et al.* RoBERTa: A robustly optimized BERT pretraining approach. arXiv [cs.CL]. 2019.http://arxiv.org/abs/1907.11692

41  Yang X, Chen A, PourNejatian N, *et al.* A large language model for electronic health records. *NPJ Digit Med* 2022;**5**:194.

42  Raffel C, Shazeer N, Roberts A, *et al.* Exploring the limits of transfer learning with a unified text-to-text transformer. arXiv [cs.LG]. 2019.http://arxiv.org/abs/1910.10683